\documentclass[10pt,twocolumn,letterpaper]{article}

\usepackage{wacv}
\usepackage{times}
\usepackage{epsfig}
\usepackage{graphicx}
\usepackage{amsmath}
\usepackage{amssymb}
\usepackage{booktabs}
\usepackage{amsfonts}
\usepackage{mathtools}
\usepackage{xcolor}
\usepackage{multirow}
\usepackage{float}
\usepackage{subcaption}
\usepackage{pifont}
\newcommand{\cmark}{\ding{51}}%
\newcommand{\xmark}{\ding{55}}%
\usepackage{makecell}

\def\wacvPaperID{624} 

\wacvalgorithmstrack   

\wacvfinalcopy 

\ifwacvfinal
\usepackage[breaklinks=true,bookmarks=false]{hyperref}
\else
\usepackage[pagebackref=true,breaklinks=true,colorlinks,bookmarks=false]{hyperref}
\fi

\def\eg{\emph{e.g.}~}

\def\etal{\emph{et al.}~}
\def\ie{\emph{i.e.}~}

\def\inp{\text{in}}
\def\outp{\text{out}}

 \newcommand{\hl}[1]{#1}

\begin{document}

\title{Uplift and Upsample: Efficient 3D Human Pose Estimation with Uplifting Transformers}

\author{Moritz Einfalt \hspace{1.7cm} Katja Ludwig \hspace{1.7cm} Rainer Lienhart\\
Machine Learning and Computer Vision Lab, University of Augsburg\\
{\tt\small \{moritz.einfalt, katja.ludwig, rainer.lienhart\}@uni-a.de}
}

\maketitle
\thispagestyle{empty}

\begin{abstract}
  The state-of-the-art for monocular 3D human pose estimation in videos is dominated by the paradigm of 2D-to-3D pose uplifting.
  While the uplifting methods themselves are rather efficient, the true computational complexity
  depends on the per-frame 2D pose estimation.
  In this paper, we present a Transformer-based pose uplifting scheme that can operate on temporally sparse 2D pose sequences but still produce temporally dense 3D pose estimates.
  We show how masked token modeling can be utilized for temporal upsampling within Transformer blocks.
  This allows to decouple the sampling rate of input 2D poses and the target frame rate of the video and drastically decreases the total computational complexity.
  Additionally, we explore the option of pre-training on large motion capture archives, which has been largely neglected so far.
  We evaluate our method on two popular benchmark datasets: Human3.6M and MPI-INF-3DHP.
  With an MPJPE of $45.0$~mm and $46.9$~mm, respectively, our proposed method can compete with the state-of-the-art while reducing inference time by a factor of 12.
  This enables real-time throughput with variable consumer hardware in stationary and mobile applications.
  We release our code and models at \url{https://github.com/goldbricklemon/uplift-upsample-3dhpe}
\end{abstract}


\section{Introduction}
\label{sec:intro}

\begin{figure}[t]
  \begin{centering}
    \includegraphics[width=\columnwidth]{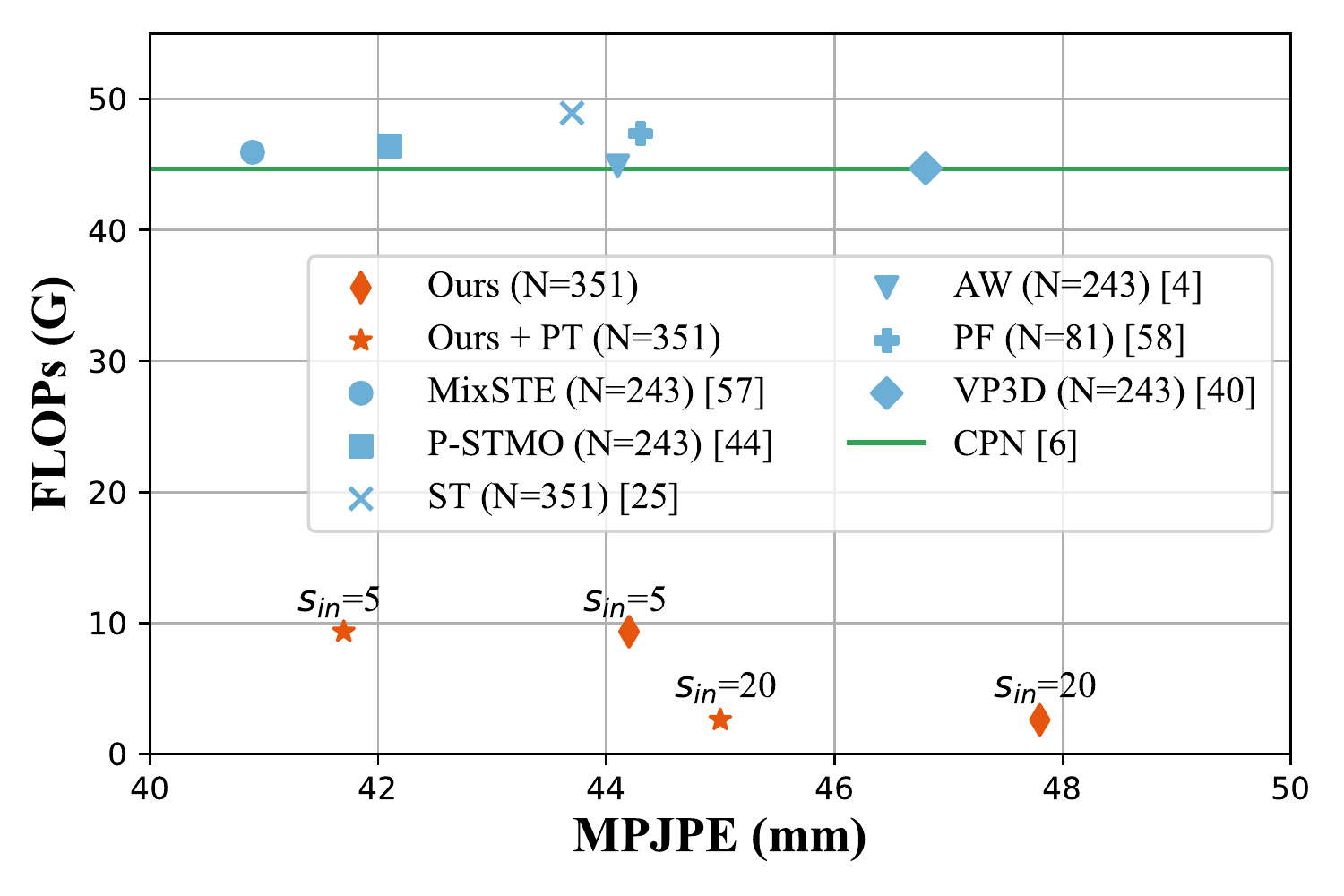}
    \caption{Spatial precision (MPJPE) and per-frame computational complexity (FLOPs) for different pose uplifting methods on Human3.6M (lower is better). The measured FLOPs include the necessary 2D pose estimation, here with CPN~\cite{chen2018cascaded}. \emph{+PT} denotes pre-training on motion capture data.}
    \label{fig:flops}
  \end{centering}
\end{figure}

\begin{figure*}[t]
  \begin{centering}
    \includegraphics[width=0.8\linewidth]{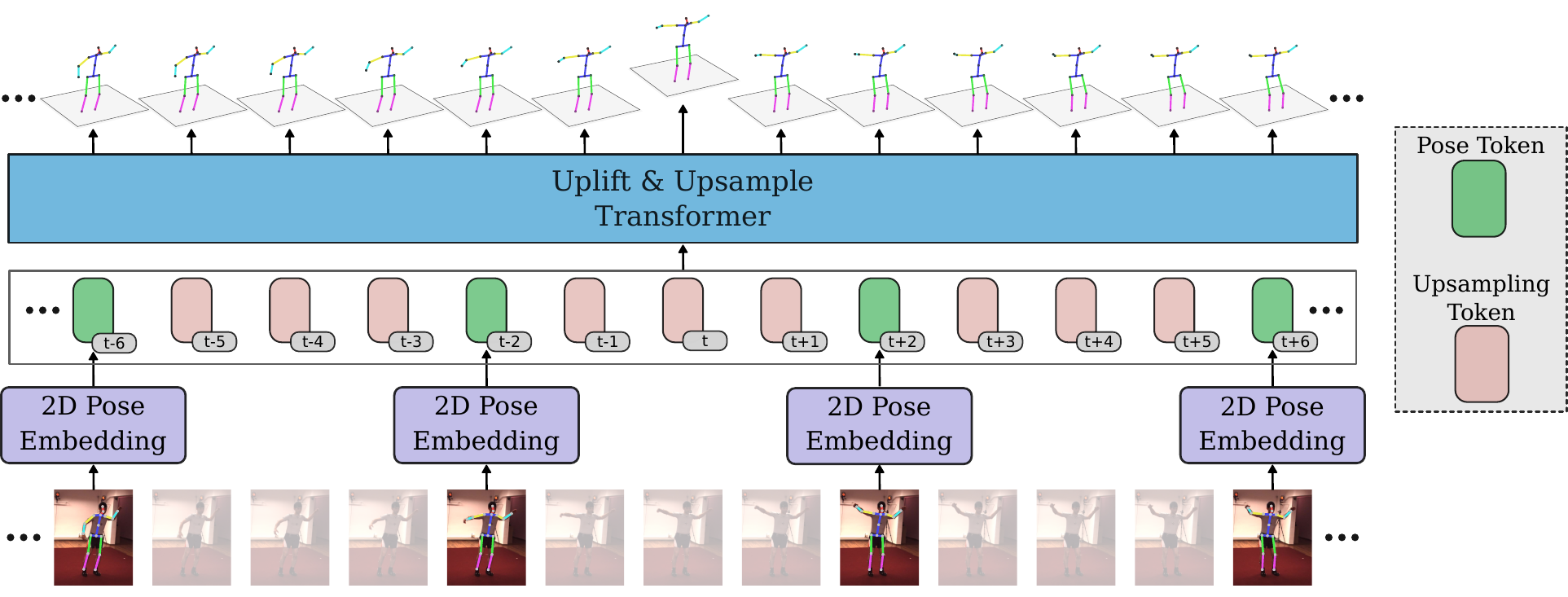}
    \caption{We extract 2D poses at a fixed key-frame interval and transform them into pose tokens. After padding this sequence with a learnable upsampling token, our Transformer network generates dense 3D pose predictions at the target frame rate. During inference, we only use the prediction for the central frame and the entire video is processed in sliding-window fashion.}
    \label{fig:teaser}
  \end{centering}
\end{figure*}

Following the huge advancements in 2D human pose estimation (HPE) over the last years, much research has been dedicated to monocular 3D HPE.
Reconstructing the locations of human joints or other body landmarks in 3D space from a single RGB camera has huge potential, with applications in computer animation~\cite{mehta17_2, mehta2020xnect, pullen2002motion}, action recognition~\cite{luvizon18, yan2018spatial, liu2017robust, li2021memory, chen2021learning},
or posture and motion analysis ~\cite{Wu2020, prima2019, wang2019ai}.
Monocular 3D HPE becomes even more relevant when the complete process can be handled on mobile computers or smartphones.
It opens up yet another field of control and interaction applications~\cite{garcia2019human, hwang2020lightweight}.

\hl{Current methods for 3D HPE in videos mainly follow the paradigm of 2D-to-3D pose uplifting~\cite{martinez2017, drover18, wandt19}.}
\hl{This two-stage approach consistently leads to the highest spatial precision on common 3D HPE benchmarks~\cite{ionescu14, mehta17}.}
It utilizes an existing image-based 2D HPE model to generate 2D poses for every single video frame.
Then, a separate uplifting model is trained to estimate the 3D poses for a sequence of frames based on only the sequential 2D pose estimates~\cite{cai2019, pavllo19, lin2019, xu2020, wang20, tripathi2020posenet3d, hu2021conditional}.
\hl{Since the 2D poses are a rather compact input description, the uplifting paradigm allows for models that operate on very long input sequences that can span multiple seconds in high frame rate recordings ~\cite{shan2021improving, chen2021anatomy, li2022exploiting, li22mhformer}.}
\hl{This is otherwise hardly possible when directly operating on the raw video frames~\cite{kanazawa19, kocabas2020}.}
Ongoing research mainly focuses on further improving the spatial precision of 3D pose estimates.
Some recent work also analyzes the computational complexity of the uplifting process itself, but fails to capture the true complexity \hl{for possible applications~\cite{zheng20213d, shan22, zhang22}.}
This comes from the dependence on 2D HPE models with high spatial precision, such as Mask R-CNN~\cite{he2017mask}, CPN~\cite{chen2018cascaded} or HRNet~\cite{cheng2020higherhrnet}.
The total complexity of 3D HPE via pose uplifting is typically dominated by the initial 2D pose estimation process (see Figure~\ref{fig:flops}). This prohibits applications where real-time throughput on compute- and power-limited consumer hardware is required.

In this paper, our main goal is to reduce the overall complexity by limiting the 2D pose estimation to a small fraction of video frames.
Existing uplifting models on 2D pose sequences always have the same rate of input and \hl{output poses~\cite{pavllo19, chen2021anatomy, zheng20213d}.}
In contrast, we present a Transformer-based architecture that can operate on temporally sparse input poses but still generate dense 3D pose sequences at the target frame rate.
Inspired by the masked token modeling in Transformer architectures~\cite{devlin2018, su2019vl, lin2021end, shan22}, we present a tokenization mechanism that allows to upsample the temporal sequence representation within our uplifting Transformer.
Missing 2D pose estimates in the input sequence are replaced by position-aware upsampling tokens.
These tokens are jointly transformed into 3D pose estimates for their respective video frames via self-attention over the entire sequence (see Figure~\ref{fig:teaser}).
This greatly reduces the computational complexity and makes our model more flexible regarding the effective input frame rate in potential applications. In fact, the sampling rate of 2D poses can even be adapted based on the expected or observed motion speed.
Since training only require annotated 3D pose sequences but no video recordings, we additionally explore pre-training on large-scale motion capture archives.
We evaluate its benefits for our and existing Transformer architectures and show how it can counter the adverse effects of sparse input sequences. In summary, our contributions are: (1)~We propose a joint uplifting and upsampling Transformer architecture that can generate temporally dense 3D pose predictions from sparse sequences of 2D poses; (2)~We evaluate the effect of Transformer pre-training on motion capture data; (3)~We show that our method leads to smoother and more precise 3D pose estimates than direct interpolation on sparse output sequences from competing methods.
At the same time, it reduces inference time by at least factor $12$ and supports different input frame rates during inference.
To the best of our knowledge, this is the first paper to explicitly address efficient 2D-to-3D pose uplifting in videos with a sparse-to-dense prediction scheme as well as direct pre-training on large-scale motion capture data.

\section{Related Work}
\label{sec:related_work}

\paragraph{2D-to-3D Pose Uplifting in Videos}
Recent work on 2D-to-3D pose uplifting in videos either uses temporal convolutional networks (TCN)~\cite{bai2018empirical}, graph convolutional networks (GCN)~\cite{defferrard2016convolutional} or Transformer networks~\cite{vaswani2017attention}.
Pavllo~\etal\cite{pavllo19} introduce a TCN-based uplifting model that can leverage long input sequences and partially labeled data.
Extensions for this model focus on attention mechanisms~\cite{liu20} or different pose representations~\cite{xu2020, chen2021anatomy}.
Cai~\etal\cite{cai2019} propose a GCN-based method that explicitly models dependencies between locally related human joints over space and time and performs well on short input sequences.
This framework is extended with either fixed~\cite{wang20} or input-conditioned non-local dependencies~\cite{hu2021conditional}.
Most recently, Transformer architectures from the vision domain are adapted for pose uplifting.
Zheng~\etal\cite{zheng20213d} introduce a Transformer architecture for temporal uplifting with self-attention over space or time.
Li~\etal\cite{li2022exploiting} use a strided Transformer block to handle longer input sequences more efficiently.
Zhang~\etal\cite{zhang22} propose a joint-wise temporal attention block for smaller and more efficient uplifting Transformers.
We follow the trend of Transformer-based uplifting and use building blocks from \cite{zheng20213d, li2022exploiting} to form a deeper architecture that is designed for sparse input sequences.

\paragraph{Efficient 3D HPE}
Most existing work on efficient 3D HPE focuses on end-to-end vision models for single images.
Mehta~\etal\cite{mehta17_2} propose a custom
CNN with real-time capabilities on consumer GPUs.
Hwang~\etal\cite{hwang2020lightweight} use CNN model compression to learn a very efficient model for mobile devices.
These methods lack reasoning over multiple video frames, which is crucial for countering the 2D-to-3D ambiguity of monocular 3D pose reconstruction~\cite{li2022exploiting, zhang22}.
Mehta~\etal\cite{mehta2020xnect} propose a real-time capable mixed CNN/MLP architecture to predict single-frame 3D poses
with additional temporal skeleton fitting.
In the scope of temporal 2D-to-3D pose uplifting, most methods are rather efficient with real-time speed on consumer hardware~\cite{pavllo19, shan22, zhang22}.
This, however, does not factor in the computational requirements for the initial per-frame 2D pose estimation. In combination, the uplifting itself accounts for only a fraction of the total complexity.
We address this issue and show that pose uplifting on sparse input sequences can easily reach real-time throughput (\ie with some constant delay) while maintaining spatial prediction accuracy.

\paragraph{Data Augmentation and Pre-Training}
3D HPE has the common difficulty that datasets with paired video and 3D pose data are scarce and have limited variability in visual appearance and human motion.
2D-to-3D pose uplifting methods have the advantage that paired data is not required~\cite{drover18, wandt19}.
This allows for data augmentation strategies on 3D motion sequences alone. Li~\etal\cite{Li2020Cascaded} use evolutionary operators to generate variations of existing 3D poses.
Gong~\etal\cite{gong2021poseaug} use a 
generative model to create plausible new 3D poses.
Both methods are limited to single poses, however.
Gong~\etal\cite{gong2022posetriplet} use a hallucination model to predict a new motion sequence from a given start and end pose, which is then refined by a dynamic skeleton model within a physics simulation.
In the scope of pre-training, Transformer architectures are known to benefit from training on large-scale datasets, including Transformers in the vision domain~\cite{carion2020end, dosovitskiy2021an, touvron2021training}.
Shan~\etal\cite{shan22} show how an uplifting Transformer can be trained with the pre-text task of 2D pose sequence reconstruction.
In contrast, we evaluate the benefits of pre-training on archives of unpaired motion capture data, for our and existing Transformer architectures.

\begin{figure*}[t]
  \begin{centering}
    \includegraphics[width=1\linewidth]{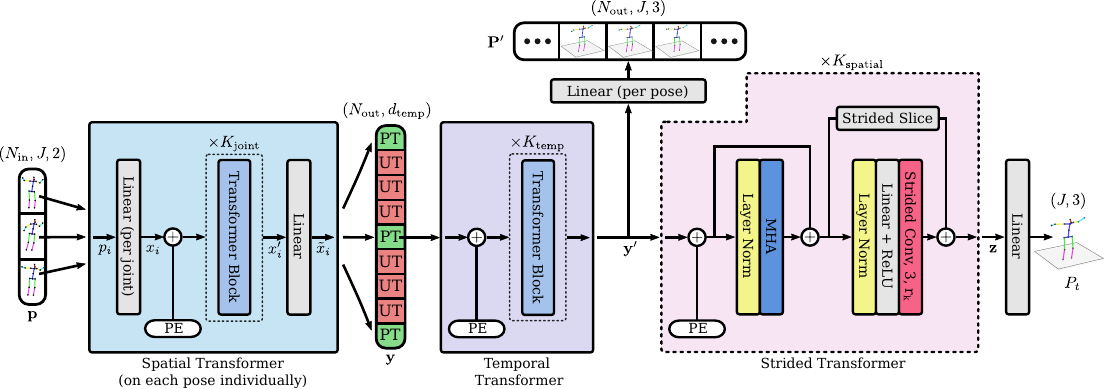}
    \caption{Instantiation of our architecture, with $N=9$, input stride $s_\inp =4$ and output stride $s_\outp =1$. The spatial and temporal Transformer use a learnable positional embedding (PE) and  vanilla Transformer blocks with multi-head self-attention (MHA). The strided Transformer employs strided convolution with stride $r_k$ and kernel size $3$. PT denotes a pose token $\tilde{x}_i$, UT the learnable upsampling token.}
    \label{fig:architecture}
  \end{centering}
\end{figure*}

\section{Method}
\label{sec:method}

Our method follows the common 2D-to-3D pose uplifting approach:
Given a video recording $V$ with frame indices $\mathcal{S}=\lbrace 1,\dotsc,|V| \rbrace$, we use an off-the-shelf 2D HPE model to obtain 2D input poses $\lbrace p_t \rbrace_{t \in \mathcal{S}}$.\footnote {We use this notation to describe an ordered sequence.}
Each pose $p_t \in \mathbb{R}^{J \times 2}$ is described by the normalized 2D image coordinates~\footnote {We assume known camera intrinsics and map 2D coordinates to $[-1, 1]$ while preserving the aspect ratio.} of $J$ designated human joints.
\hl{We discard the detection scores or any other occlusion information that the 2D HPE model might provide.}
The goal is to recover the actual 3D pose $P_t^{\text{gt}} \in \mathbb{R}^{J \times 3}$, \ie the metric camera coordinates of every joint, for every single video frame.
Our main focus lies on using only a subset of input 2D poses at equidistant key-frames $\mathcal{S}_\inp = \lbrace t | t \bmod s_\inp = 0 \rbrace_{t \in \mathcal{S}}$ with input stride $s_\inp$. At the same time, we want to generate 3D pose predictions at a higher rate, \ie with a smaller output stride $s_\outp < s_\inp$, at frame indices $\mathcal{S}_\outp = \lbrace t | t \bmod s_\outp = 0 \rbrace_{t \in \mathcal{S}}$. For simplicity, we will assume $s_\inp = k \cdot s_\outp$, $k \in \mathbb{N}$, such that $\mathcal{S}_\inp \subseteq \mathcal{S}_\outp$. Ideally, with $s_\outp = 1$, we can predict dense 3D poses at the full frame rate.
\hl{Figure~\ref{fig:teaser} depicts an example with $\mathcal{S}_\inp=\lbrace 1,5,9,\dotsc \rbrace$ and $\mathcal{S}_\outp=\lbrace 1,2,3,\dotsc \rbrace$.}

\subsection{Uncoupling of Input and Output Sample Rate}

Existing uplifting methods have an identical input/output sample rate of 2D/3D \hl{poses~\cite{pavllo19, chen2021anatomy, zheng20213d}.}
Each model is optimized for 
a fixed input rate. When trained on sub-sampled input poses with stride $s_\inp > 1$, the model can only predict 3D poses at an equal output stride $s_\outp = s_\inp$.
To obtain predictions at the full frame rate, we can use bilinear interpolation between 
3D poses at adjacent key-frames.
This naive method has two drawbacks: First, each model is still optimized for a single, now reduced, frame rate.
This lacks flexibility regarding applications with different input frame rates or variable computational resources.
Second, using large values for $s_\inp$ will increasingly deteriorate the predictions at the full frame rate. Simple bilinear interpolation cannot reconstruct the actual human motion between two key-frames.
Next, we describe our proposed architecture that can circumvent both issues by uncoupling $s_\inp$ and $s_\outp$.

\subsection{Joint Uplifting and Upsampling}

Our solution follows the recent trend of 2D pose sequence uplifting with Transformer \hl{networks~\cite{zheng20213d}.} The self-attentive nature of Transformers has shown to be well suited to leverage relationships between individual joints of a single human pose~\cite{Li_2021_TokenPose} and within long sequences of poses in time~\cite{li2022exploiting, zhang22}.
The main design criterion for our uplifting network is a comparatively deeper architecture that can operate on variably sparse input sequences. It has to concurrently handle pose uplifting and temporal upsampling. At the same time, the model should stay efficient in training and inference.
Similar to most \hl{temporal} pose uplifting methods, \hl{our model operates on 2D pose sequences with fixed length $N=2n+1$, $n \in \mathbb{N}$. Each sequence covers frame indices $\mathcal{S}_N=\lbrace t-n, \dotsc, t+n \rbrace$ around an output frame $t \in \mathcal{S}_\outp$.
Since our Transformer-based model aggregates information from the entire sequence, its temporal receptive field~\cite{pavllo19} equals $N$.
However, due to key-frame sub-sampling, the actual input to our model is comprised of only the key-frame 2D poses $\mathbf{p}$ with:}
\begin{equation}
  \label{eq:input_seq}
  \mathbf{p} = \lbrace p_i | i \in \mathcal{S}_\inp \cap \mathcal{S}_N  \rbrace.
\end{equation}
The effective input sequence length is therefore reduced to $N_\inp \vcentcolon = |\mathbf{p}| \leq \frac{N-1}{s_\inp}+1$. During training, the model generates intermediate 3D pose predictions $P_i^{\prime} \in \mathbb{R}^{J \times 3}$ for all output frames within the temporal receptive field:
\begin{equation}
  \label{eq:output_seq}
  \mathbf{P}^\prime = \lbrace P_i^\prime | i \in \mathcal{S}_\outp \cap \mathcal{S}_N  \rbrace.
\end{equation}
Additionally, the 3D pose for the central frame $t$ is further refined to its final prediction $P_t$. During evaluation, only this central prediction is used. We utilize three distinct Transformer sub-networks from recent literature that, in combination, fit our main design goals. Figure~\ref{fig:architecture} provides an overview of our architecture.

\paragraph{Joint-Wise Spatial Transformer}

The first sub-network is shared between all input poses and operates on each pose individually.
It uses self-attention across individual joints to form a strong pose representation for the subsequent sub-networks.
Each input pose $p_i$ is first linearly mapped to an initial joint-wise embedding $x_i \in \mathbb{R}^{J \times d_{\text{joint}}}$. After adding a positional embedding to encode the type of each joint, we use $K_{\text{joint}}$ spatial Transformer blocks~\cite{zheng20213d} that operate on the sequence of joint embeddings. The output is the joint-aware per-pose embedding $x_i^\prime \in \mathbb{R}^{J \times d_{\text{joint}}}$. It is subsequently condensed into a 1D encoding $\tilde{x}_i \in \mathbb{R}^{d_{\text{temp}}}$. We refer to all $\tilde{x}_i$ as our initial pose tokens.

\paragraph{Pose-Wise Temporal Transformer}
The second sub-network uses vanilla Transformer blocks with self-attention across the temporal sequence of pose tokens.
This building block is a quasi-standard in recent Transformer-based uplifting methods~\cite{zheng20213d, li2022exploiting, shan22}.
We extend its usual objective of direct 3D pose reconstruction for key-frame poses within the input sequence.
We want to generate smooth and temporally consistent 3D poses for all output frames within the temporal receptive field.
We present a simple modification that enables simultaneous uplifting and upsampling within the temporal Transformer blocks.
First, we recombine the key-frame pose tokens  $\tilde{x}_i$ from the spatial Transformer into a temporal sequence.
We then pad this sequence to target length $N_{\outp} \vcentcolon = |\mathbf{P}^\prime|$.
For this, we adopt the masked token modeling of Transformers~\cite{lin2021end, shan22} and introduce an upsampling token $u \in \mathbb{R}^{d_{\text{temp}}}$.
\hl{It is a learnable parameter that is initialized randomly and optimized during training.}
This token acts as a placeholder at all non-key-frame indices.
Figure~\ref{fig:teaser} depicts this gap-filling process.
The elements of the padded token sequence $\mathbf{y} = \lbrace y_i | i \in \mathcal{S}_\outp \cap \mathcal{S}_N \rbrace$ are then defined as
\begin{equation}
  \label{eq:padded_sequence}
  y_i =
  \begin{cases}
    \tilde{x}_i  &  \text{if $i \in  \mathcal{S}_\inp$}, \\
    u & \text{else}. 
  \end{cases}
\end{equation}
In contrast to~\cite{shan22}, the token $u$ not only encodes a pre-text task of input reconstruction, but rather the reconstruction of the upsampled sequence in output space.
A second positional embedding ensures that, in particular, each instantiation of the upsampling token is conditioned on its relative frame index.
We feed the token sequence to a stack of $K_{\text{temp}}$ vanilla Transformer blocks.  We restrict any attention within the first Transformer block to the pose tokens, since the initial upsampling tokens do not carry input related information.
\hl{This is implemented by only computing self-attention keys and values~\cite{vaswani2017attention} from the pose tokens.}
The output $\mathbf{y}^{\prime} \in \mathbb{R}^{N_\outp \times d_{\text{temp}}}$ after the last Transformer block encodes the uplifted and upsampled 3D poses for all output frames.
We use a single linear mapping to obtain the intermediate 3D pose predictions $\mathbf{P}^\prime$.

\paragraph{Sequence Reduction for Single-Frame Output}
One main drawback of vanilla Transformer blocks is the quadratic complexity w.r.t. the sequence length.
Stacking a large number of vanilla Transformer blocks at full temporal resolution does not align with our goal of an overall efficient model. Ultimately, our model is designed for a symmetric $N$-to-$1$ prediction scheme during evaluation. This operation mode commonly delivers superior results, as the prediction for the central sequence index is based on an equal amount of past and future information~\cite{zheng20213d, shan22}. 
To further refine the pose prediction specifically for the central index $t$, it is not necessary to keep the full sequence representation of length $N_\outp$. Our third sub-network therefore incrementally reduces the previous sequence representation $\mathbf{y}^{\prime}$, until only the refined output for the central sequence index remains.
This allows us to add additional temporal self-attention blocks, but keep the overall complexity feasible.
We utilize $K_{\text{stride}}$ strided Transformer blocks~\cite{li2022exploiting}, which use strided convolutions instead of a simple MLP.
The details are depicted in Figure~{\ref{fig:architecture}}.
Each block $k$ reduces the sequence length by stride factor $r_k$.
We choose all $r_k$ such that the output $\mathbf{z}$ after the last block is $\mathbf{z}\in \mathbb{R}^{1 \times d_{\text{temp}}}$. A single linear mapping generates the final 3D pose prediction $P_{t} \in \mathbb{R}^{J \times 3}$ for the central sequence index.


\paragraph{Sequence and Center Supervision}
The entire architecture is trained with two separate objectives.
We use the \hl{center frame loss}  $\mathcal{L}_{\text{center}}$ to minimize the root-relative mean per-joint position error (MPJPE)~\cite{ionescu14} of the refined 3D pose prediction for \hl{central} target frame $t$:
\begin{equation}
  \label{eq:center_loss}
  \mathcal{L}_{\text{center}} = \frac{1}{J} \sum_{j=1}^{J} \lVert (P_{t,j} - P_{t, r}) - (P^{\text{gt}}_{t,j} - P^{\text{gt}}_{t, r}) \rVert_2,
\end{equation}
where the pelvis is commonly used as the designated root joint $r$.
Additionally, we define the MPJPE sequence loss $\mathcal{L}_{\text{seq}}$ on the intermediate 3D pose predictions $P_i^{\prime}$ for the entire upsampled sequence:
\begin{equation}
  \label{eq:sequence-loss}
  \mathcal{L}_{\text{seq}} = \frac{1}{J \cdot N_\outp} \sum_{i \in \mathcal{S}_\outp \cap \mathcal{S}_N} \sum_{j=1}^{J} \lVert (P^{\prime}_{i,j} - P^{\prime}_{i, r}) - (P^{\text{gt}}_{i,j} - P^{\text{gt}}_{i, r}) \rVert_2.
\end{equation}
 This form of full-sequence supervision encourages temporally stable predictions~\cite{li2022exploiting, zhang22}, which is especially important in our setting of sparse input sequences.
The total loss is the weighted sum $\alpha_1 \mathcal{L}_{\text{seq}} + \alpha_2 \mathcal{L}_{\text{center}}$.


\paragraph{Training and Inference}
We optimize our model for a fixed output stride $s_\outp$, but for multiple input strides $s_\inp$ simultaneously.
Thus, it supports different input frame rates, depending on the application and the available hardware.
During training, we utilize 3D pose annotations at the full frame rate and generate all \hl{possible} key-frame sequences around each frame index $t \in \mathcal{S}$.
For inference, only the key-frame poses starting at \hl{the first video frame} are available and we apply our model at every output frame $t \in \mathcal{S}_\outp$.
In case of $s_\outp > 1$, 3D pose predictions at the full frame rate are obtained via bilinear interpolation. We always evaluate at the full video frame rate for fair comparison.

\subsection{MoCap Pre-Training}
In order to further unlock the potential of Transformer architectures in 2D-to-3D pose uplifting, we additionally explore the effects of pre-training on large-scale motion capture data.
In this work, we utilize the AMASS~\cite{Mahmood_2019_ICCV_AMASS} meta-dataset.
It is a collection of a wide variety of existing motion capture datasets with over $60$ hours of human motion.
The raw motion capture data has been re-targeted to generate a detailed 3D mesh model of the captured person in compact SMPL parameterization~\cite{loper2015smpl}.
\hl{We reduce the mesh to the target set of $J$ joints.
Each joint's 3D location is expressed as a weighted linear combination of a small set of mesh vertices.
The mixing weights can either be directly optimized on data~\cite{kolotouros2019SPIN} or created by a small number of hand annotations.
Finally, we project the 3D pose sequences to 2D space with randomly selected virtual cameras.
For simplicity, we use the same camera parameters from our final target datasets.
The resulting 2D-3D pose sequence pairs can then be directly used for training.
Note that the 2D poses are perfect projections and thus without errors.
Our model will adjust to the error cases of a 2D pose estimation model during subsequent fine-tuning.}

\section{Experiments}
\label{sec:experiments}

\begin{table*}[t]\small
  \begin{center}
    \caption{Results on Human3.6M with \hl{CPN~\cite{chen2018cascaded}} 2D poses. We evaluate according to Protocol 1 (MPJPE, top) and Protocol 2 (P-MPJPE, bottom). Best results are bold, second best results are underlined. (*) uses the refinement module from~\cite{cai2019}. \textit{+PT} denotes MoCap pre-training on AMASS.}
    \label{tab:h36m_results}
    \resizebox{\linewidth}{!}{
      \begin{tabular}{l|ccccccccccccccc|c}
        \hline
        \noalign{\smallskip}
        MPJPE (mm)$\downarrow$ & Dir. & Disc. & Eat & Greet & Phone & Photo & Pose & Pur. & Sit & SitD. & Smoke & Wait & WalkD. & Walk & WalkT. & Avg \\
        \noalign{\smallskip}
        \hline
        \noalign{\smallskip}
        Cai \etal\cite{cai2019} ICCV'19 ($N$ = 7)(*)& 44.6& 47.4& 45.6& 48.8& 50.8& 59.0& 47.2& 43.9& 57.9& 61.9& 49.7& 46.6& 51.3& 37.1& 39.4& 48.8\\
        Pavllo \etal\cite{pavllo19} CVPR'19 ($N$ = 243)& 45.2&46.7&43.3&45.6&48.1&55.1&44.6&44.3&57.3&65.8&47.1&44.0&49.0&32.8&33.9&46.8 \\
        Xu \etal\cite{xu2020} CVPR'20 ($N$ = 9)
              &\textbf{37.4}&43.5&42.7&42.7&46.6&59.7&41.3&45.1&\underline{52.7}&60.2&45.8&43.1&47.7&33.7&37.1&45.6\\
        Zheng \etal\cite{zheng20213d} ICCV'21 ($N$ = 81)&
                                                         41.5&44.8&39.8&42.5&46.5&51.6&42.1&42.0&53.3&60.7&45.5&43.3&46.1&31.8&32.2&44.3\\
        Shan \etal\cite{shan2021improving} MM'21 ($N$ = 243)&
                                                             40.8&44.5&41.4&42.7&46.3&55.6&41.8&41.9&53.7&60.8&45.0&41.5&44.8&30.8&31.9&44.3\\
        Chen \etal\cite{chen2021anatomy} TCSVT'21 ($N$ = 243)&
                                                              41.4&43.5&40.1&42.9&46.6&51.9&41.7&42.3&53.9&60.2&45.4&41.7&46.0&31.5&32.7&44.1\\
        Li \etal\cite{li2022exploiting} TMM'22 ($N$ = 351)(*)&
                                                                40.3&43.3&40.2&42.3&45.6&52.3&41.8&40.5&55.9&60.6&44.2&43.0&44.2&30.0&30.2&43.7\\
        Hu \etal\cite{hu2021conditional} MM'21 ($N$ = 96)&
                                                           38.0&43.3&39.1&\textbf{39.4}&45.8&53.6&41.4&41.4&55.5&61.9&44.6&41.9&44.5&31.6&29.4&43.4\\
        Li \etal\cite{li22mhformer} CVPR'22 ($N$ = 351)& 39.2&43.1&40.1&40.9&44.9&51.2&40.6&41.3&53.5&60.3&43.7&41.1&43.8&29.8&30.6& 43.0\\
        Shan \etal\cite{shan22} arXiv'22 ($N$ = 243)(*)  &
                                                          38.4& 42.1&39.8&40.2&45.2&48.9&\underline{40.4}&\textbf{38.3}&53.8&\underline{57.3}&43.9&41.6&42.2&29.3&29.3&42.1\\
        Zhang \etal\cite{zhang22} CVPR'22 ($N$ = 243)& \underline{37.6}& \textbf{40.9}& \textbf{37.3}& \underline{39.7}& \textbf{42.3}& 49.9& \textbf{40.1}& \underline{39.8}& \textbf{51.7}& \textbf{55.0}& \textbf{42.1}& \textbf{39.8}& \textbf{41.0}& \textbf{27.9}& \textbf{27.9}& \textbf{40.9}\\
        \noalign{\smallskip}
        \hline
        \noalign{\smallskip}
        Ours ($N$ = 351), $s_\inp$ = $s_\outp$ = 5 &41.8&45.5&41.8&44.2&48.4&54.2&43.7&43.1&58.9&66.3&46.1&43.7&46.0&30.9&31.2&45.7 \\ 
        Ours ($N$ = 351), $s_\inp$ = $s_\outp$ = 5 (*) &39.6&43.8&40.2&42.4&46.5&53.9&42.3&42.5&55.7&62.3&45.1&43.0&44.7&30.1&30.8& 44.2 \\  
        Ours ($N$ = 351), $s_\inp$ = $s_\outp$ = 5, + PT &40.6&42.7&38.5&41.1&45.2&\underline{48.7}&41.5&41.0&53.3&61.3&43.3&\underline{41.0}&42.3&30.0&\underline{29.0}& 42.6 \\  
        Ours ($N$ = 351), $s_\inp$ = $s_\outp$ = 5, + PT (*) &38.6&\underline{41.0}&\underline{37.6}&\underline{39.7}&\underline{44.2}&\textbf{47.9}&40.9&\underline{39.8}&\textbf{51.7}&60.3&\underline{43.1}&41.1&\underline{41.6}&\underline{28.4}&29.2& \underline{41.7} \\ 
        Ours ($N$ = 351), $s_\inp$ = 20, $s_\outp$ = 5 &45.4&47.9&43.4&47.2&49.6&55.9&46.4&45.4&59.9&66.7&47.5&45.5&49.8&33.0&33.8&47.8 \\ 
        Ours ($N$ = 351), $s_\inp$ = 20, $s_\outp$ = 5, + PT &44.5&45.1&40.3&44.6&46.3&50.7&44.4&43.7&54.6&62.3&44.9&43.1&47.0&32.3&31.9& 45.0  \\
        \noalign{\smallskip}
        \hline
        \hline
        \noalign{\smallskip}
        P-MPJPE (mm)$\downarrow$ & Dir. & Disc. & Eat & Greet & Phone & Photo & Pose & Pur. & Sit & SitD. & Smoke & Wait & WalkD. & Walk & WalkT. & Avg \\
        \noalign{\smallskip}
        \hline
        \noalign{\smallskip}
        Cai \etal\cite{cai2019} ICCV'19 ($N$ = 7)(*)&35.7 &37.8 &36.9 &40.7 &39.6 &45.2 &37.4 &34.5 &46.9 &50.1 &40.5 &36.1 &41.0 &29.6 &33.2 &39.0 \\
        Pavllo \etal\cite{pavllo19} CVPR'19 ($N$ = 243)& 34.1&36.1&34.4&37.2&36.4&42.2&34.4&33.6&45.0&52.5&37.4&33.8&37.8&25.6&27.3&36.5 \\
        Xu \etal\cite{xu2020} CVPR'20 ($N$ = 9)
              &31.0&34.8&34.7&34.4&36.2&43.9&31.6&33.5&\underline{42.3}&49.0&37.1&33.0&39.1&26.9&31.9&36.2\\
        Chen \etal\cite{chen2021anatomy} TCSVT'21 ($N$ = 243)&
                                                              33.1&35.3&33.4&35.9&36.1&41.7&32.8&33.3&42.6&49.4&37.0&32.7&36.5&25.5&27.9&35.6\\
        Li \etal\cite{li2022exploiting} TMM'22 ($N$ = 351)(*)&32.7 &35.5 &32.5 &35.4 &35.9 &41.6 &33.0 &31.9 &45.1 &50.1 &36.3 &33.5 &35.1 &23.9 &25.0 &35.2   \\
        Li \etal\cite{li22mhformer} CVPR'22 ($N$ = 351)&31.5&34.9&32.8&33.6&35.3&39.6&32.0&32.2&43.5&48.7&36.4&32.6&34.3&23.9&25.1&34.4\\
        Shan \etal\cite{shan2021improving} MM'21 ($N$ = 243)& 32.5&36.2&33.2&35.3&35.6&42.1&32.6&31.9&42.6&\underline{47.9}&36.6&32.1&34.8&24.2&25.8&35.0\\
        Zheng \etal\cite{zheng20213d} ICCV'21 ($N$ = 81)& 32.5&34.8&32.6&34.6&35.3&39.5&32.1&32.0&42.8&48.5&\underline{34.8}&32.4&35.3&24.5&26.0&34.6\\
        Shan \etal\cite{shan22} arXiv'22 ($N$ = 243)    &31.3&35.2&32.9&33.9&35.4&39.3&32.5&31.5&44.6&48.2&36.3&32.9&34.4&23.8&23.9&34.4\\
        Hu \etal\cite{hu2021conditional} MM'21 ($N$ = 96)&\textbf{29.8} &34.4 &31.9 &\textbf{31.5} &35.1 &40.0 &\textbf{30.3} &\underline{30.8} &42.6 &49.0 &35.9 &\underline{31.8} &35.0 &25.7 &\underline{23.6} &\underline{33.8} \\
        Zhang \etal\cite{zhang22} CVPR'22 ($N$ = 243)&\underline{30.8} &\textbf{33.1} &\textbf{30.3} &\underline{31.8} &\textbf{33.1} &\underline{39.1} &\underline{31.1} &\textbf{30.5} &42.5 &\textbf{44.5} &\textbf{34.0} &\textbf{30.8} &\textbf{32.7} &\textbf{22.1} &\textbf{22.9} &\textbf{32.6}  \\

        \noalign{\smallskip}
        \hline
        \noalign{\smallskip}
        Ours ($N$ = 351), $s_\inp$ = $s_\outp$ = 5 (*) &32.7&36.1&33.4&36.0&36.1&42.0&33.3&33.1&45.4&50.7&37.0&34.1&35.9&24.4&25.4& 35.7 \\  
        Ours ($N$ = 351), $s_\inp$ = $s_\outp$ = 5, + PT (*) &31.6&\underline{33.7}&\underline{31.8}&33.3&\underline{34.7}&\textbf{38.7}&32.2&31.2&\textbf{41.9}&48.9&35.5&32.6&\underline{33.7}&\underline{23.4}&24.0&  \underline{33.8} \\ 
        Ours ($N$ = 351), $s_\inp$ = 20, $s_\outp$ = 5 &37.6&38.9&36.0&39.4&38.2&44.1&36.4&35.2&48.3&52.9&38.6&35.8&39.6&26.8&27.5&38.4 \\ 
        Ours ($N$ = 351), $s_\inp$ = 20, $s_\outp$ = 5, + PT &36.4&36.5&33.0&37.1&36.4&40.1&34.8&34.3&45.0&50.1&36.9&34.2&37.9&26.5&25.8&  36.3 \\
        \noalign{\smallskip}
        \hline

      \end{tabular}
    }
  \end{center}
\end{table*}

We evaluate our proposed method on two well-known 3D HPE datasets and compare it with the current state-of-the-art in 2D-to-3D pose uplifting.
We also conduct a series of ablation experiments to reveal the impact of sparse input sequences, explicit upsampling and large-scale pre-training on spatial accuracy and inference efficiency.
Additional experiments on multi-stride training, augmentation strategies and architecture components as well as qualitative examples can be found in the supplementary material.

\subsection{Datasets}
\noindent\textbf{Human3.6M}~\cite{ionescu14} is the most common dataset for indoor 3D HPE.
It consists of 11 actors performing 15 different actions each.
They are recorded by four stationary RGB cameras at $50~\text{Hz}$.
We follow the standard evaluation protocol from previous \hl{work~\cite{martinez2017, drover18, pavllo19, zheng20213d}}:
Five subjects (S1, S5, S6, S7, S8) are used for training, while evaluation is performed on two subjects (S9, S11).
We use 2D poses from a fine-tuned CPN~\cite{chen2018cascaded} during training and evaluation.\\ \\
\noindent\textbf{MPI-INF-3DHP}~\cite{mehta17} is a smaller but more challenging dataset for single-person 3D HPE with more diversity in motion, viewpoints and environments.
The training data consists of eight actors performing various actions in a green screen studio with 14 RGB cameras.
The evaluation data consists of indoor and outdoor recordings of six actors from a single camera.
We sample all recordings to a common frame rate of $25~\text{Hz}$.
Since some test-set videos are at $50~\text{Hz}$, we use additional bilinear upsampling on the estimated 3D poses to evaluate on the full frame rate.
We use ground truth 2D poses in all experiments for best comparison with existing work.\\ \\
\textbf{Metrics} We evaluate results on Human3.6M with the MPJPE \hl{metric~\cite{ionescu14}} (Equation~\ref{eq:center_loss}). We additionally report the \hl{N-MPJPE~\cite{rhodin2018geometry}} and \hl{P-MPJPE~\cite{martinez2017}}, \ie the MPJPE after scale or procrustes alignment.
\hl{Evaluation with MPJPE and P-MPJPE is often referred to as Protocol 1 and 2, respectively.}
For MPI-INF-3DHP, we report the MPJPE, the percentage of correct keypoints (PCK) at a maximum misplacement of 150~mm and the area under the curve (AUC) with a threshold range of 5-150~mm~\cite{mehta17}.

\subsection{Implementation Details}
\label{sec:implementation_details}


We instantiate our architecture with $K_{\text{joint}}=4$, $K_{\text{temp}}=4$ and $K_{\text{strided}}=3$ Transformer blocks, with an internal representation size of $d_{\text{joint}}=32$ and $d_{\text{temp}}=348$. The spatial and temporal Transformer use stochastic depth~\cite{huang2016deep} with a drop rate of $0.1$. 
We evaluate temporal receptive fields of $N \in \lbrace 81, 351 \rbrace$. For $N=351$, we use $s_\outp=5$ and train on variable input strides $s_\inp \in [5, 10, 20]$. For $N=81$, we use $s_\outp=2$ and $s_\inp \in [4, 10, 20]$.
For best results, we extend our architecture with the 3D pose refinement module from~\cite{cai2019}.
\hl{It uses camera intrinsics for reprojection to improve the orientation of some 3D pose estimates.}
Our model is trained with AdamW~\cite{loshchilov2017AdamW} for 120 epochs and a batch size of $512$.
We employ standard data augmentation with horizontal flipping of input poses during training and evaluation. Specifically, we use within-batch augmentation, where the second half of each mini-batch is the flip-augmented version of the first half.
We use an initial learning rate of $4\text{e}^{-5}$, with an exponential decay by $0.99$ per epoch.
The same schedule is applied to the initial weight decay of $4\text{e}^{-6}$. The loss weights are fixed at $\alpha_1 = \alpha_2 = 0.5$.
All experiments are conducted on a single NVIDIA A100 GPU.

\subsection{Results}

\begin{table}[t]\small
  \begin{center}
    \caption{MPI-INF-3DHP results on ground truth 2D poses.}
    \label{tab:mpiinf_results}
    \resizebox{\columnwidth}{!}{
      \begin{tabular}{l|ccc}
        \hline\noalign{\smallskip}
        Method&PCK$\uparrow$&AUC$\uparrow$&MPJPE$\downarrow$\\
        \noalign{\smallskip}
        \hline
        \noalign{\smallskip}
        Pavllo \etal\cite{pavllo19} CVPR'19 ($N$ = 81)&86.0&51.9&84.0\\
        Chen \etal\cite{chen2021anatomy} TCSVT'21 ($N$ = 81)&87.9&54.0&78.8\\
        Zheng \etal\cite{zheng20213d} ICCV'21 ($N$ = 9)&88.6&56.4&77.1\\
        Wang \etal\cite{wang20} ECCV'20 ($N$ = 96)&86.9&62.1&68.1\\
        Li \etal\cite{li22mhformer} CVPR'22 ($N$ = 9)&93.8&63.3&58.0\\
        Zhang \etal\cite{zhang22} CVPR'22 ($N$ = 27)&94.4&66.5&54.9\\
        Hu \etal\cite{hu2021conditional} MM'21 ($N$ = 96)&\textbf{97.9}&\underline{69.5}&\underline{42.5}\\
        Shan \etal\cite{shan22} arXiv'22 ($N$ = 81)&\textbf{97.9}&\textbf{75.8}&\textbf{32.2}\\
        \noalign{\smallskip}
        \hline
        \noalign{\smallskip}
        Ours ($N$ = 81), $s_\inp$ = 10, $s_\outp$ = 2 &\underline{95.4}&67.6 &46.9 \\ 
        Ours ($N$ = 81), $s_\inp$ = 10, $s_\outp$ = 2, + PT &+1.7&+2.4 &-5.7 \\ 
        \noalign{\smallskip}
        \hline
      \end{tabular}
    }
  \end{center}
\end{table}

We compare our method against recent work and the current state-of-the-art. Note that all comparative results use 2D poses at the full frame rate.
Table~\ref{tab:h36m_results} shows the results on Human3.6M.
We first evaluate our architecture with a key-frame stride of $s_\inp =5$ and no internal upsampling ($s_\inp = s_\outp$).
The full-rate 3D poses are obtained with bilinear upsampling.
Starting at a base MPJPE of $45.7$~mm, additional reprojection refinement improves spatial accuracy to $44.2$~mm.
We can see that our architecture can generate competitive results despite requiring 5 times less input poses.
When adding MoCap pre-training (\textit{+PT}), we can further improve results by \hl{$2$--$3$~mm}.
Similar behaviour can be seen for P-MPJPE results.
Since we use additional data in this experiment, we do not claim our architecture being better than existing ones.
It simply reveals that pre-training can easily compensate the reduced rate of input poses.
In order to further reduce the input rate of 2D poses for even larger efficiency gains, we make use of our joint uplifting and upsampling mechanism. With an input stride of $s_\inp = 20$, we achieve a base MPJPE of $47.8$~mm.
Reducing 2D poses to only $2.5$~Hz thus leads to an increase by $\sim 2$~mm in MPJPE.
But again, with additional pre-training, we can reduce this negative effect to a large extent and obtain competitive results with $45.0$~mm MPJPE.
At the same time, we are $20$ times more efficient in the expensive but required 2D pose estimation and only require a fifth of the forward passes for the uplifting model.

\begin{table}[t]\small
  \begin{center}
    \caption{Results on Human3.6M with $N=81$ and varying input strides $s_\inp$. Results are shown for poses on key-frames as well as all frames at $50$~Hz.}
    \label{tab:sparse_results}
    \resizebox{\columnwidth}{!}{
      \begin{tabular}{l|c|c|c}
        \hline\noalign{\smallskip}
        &&\multicolumn{2}{c}{MPJPE / N-MPJPE / P-MPJPE $\downarrow$}\\
        Method&$s_\inp$&Key-frames&All frames \\
        \noalign{\smallskip}
        \hline
        \noalign{\smallskip}
        Strided Transformer~\cite{li2022exploiting}&\multirow{4}{*}{4} & 49.3 / 47.7 / 38.7 & 49.4 / 47.8 / 38.7 \\ 
        Pose Former~\cite{zheng20213d}&  &47.7 / 46.3 /37.6& 47.7 / 46.3 / 37.6 \\ 
        \hl{Ours, $s_\outp$ = $s_\inp$} & & \textbf{47.6} / \textbf{46.0} / \textbf{37.3} & 47.7 / 46.0 / 37.4 \\ 
        Ours, $s_\outp$ = 2& & \textbf{47.6} / \textbf{46.0} / \textbf{37.3} & \textbf{47.4} / \textbf{45.8} / \textbf{37.1} \\ 
        \noalign{\smallskip}
        \hline
        \noalign{\smallskip}
        Strided Transformer~\cite{li2022exploiting}& \multirow{4}{*}{10} & 51.4 / 49.4 / 40.2 & 52.0 / 50.0 / 40.8 \\  
        Pose Former~\cite{zheng20213d}& & 48.8 / 46.9 / 38.0 & 49.3 / 47.4 / 38.5  \\ 
        \hl{Ours, $s_\outp$ = $s_\inp$} & & \textbf{47.5} / \textbf{45.8} / \textbf{37.1} & 48.1 / 46.3 / 37.6 \\ 
        Ours, $s_\outp$ = 2& & \textbf{47.5} / \textbf{45.8} / \textbf{37.1} & \textbf{47.9} / \textbf{46.1} / \textbf{37.4} \\ 
        \noalign{\smallskip}
        \hline
        \noalign{\smallskip}
        Strided Transformer~\cite{li2022exploiting}& \multirow{4}{*}{20} & 54.4 / 52.4 / 42.2 & 57.7 / 55.7 / 45.4 \\ 
        Pose Former~\cite{zheng20213d}& & 48.6 / 47.2 / 38.2 & 52.0 / 50.6 / 41.4 \\ 
        \hl{Ours, $s_\outp$ = $s_\inp$} & & \textbf{48.1} / \textbf{46.4} / \textbf{37.6} & 51.6 / 49.6 / 40.8 \\ 
        Ours, $s_\outp$ = 2& & \textbf{48.1} / \textbf{46.4} / \textbf{37.6} &  \textbf{49.9} / \textbf{48.1} / \textbf{39.2} \\ 
        \noalign{\smallskip}
        \hline
      \end{tabular}
    }
  \end{center}
\end{table}

Table~\ref{tab:mpiinf_results} shows our results on MPI-INF-3DHP. Our method is even closer to the state-of-the-art on this more challenging dataset.
With a setting of $s_\inp=10$ and recordings in $25$~Hz, we again only require input poses at $2.5$~Hz.
Despite this huge reduction in complexity, we are able to achieve the currently third best PCK, AUC and MPJPE with $95.4$, $67.6$ and $46.9$~mm, respectively.
This confirms the competitiveness of our method despite the constraint of sparse input pose sequences.
Additional MoCap pre-training leads to further improvement of $2.4$ (AUC) and $5.7$~mm (MPJPE).
Thus, independent of the target dataset, additional pre-training can reliably improve the spatial precision of 3D pose estimates from sparse 2D poses.

\subsection{Ablation Study}
We additionally explore how sparse input sequences and MoCap pre-training individually influence our architecture compared to existing uplifting Transformers with similar building blocks.
\hl{Here, we use a smaller receptive field of $N = 81$ and no refinement module~\cite{cai2019}.
For easier ablation, we adjust the training recipe of our model as well as the original recipes for the comparative methods in ~\cite{zheng20213d, li2022exploiting} by a small set of common changes: We adopt a batch size of $256$
and train for the full number of reported epochs without early stopping.
We additionally use an exponentially moving average of the model weights~\cite{tarvainen17} to reduce fluctuations in evaluation results.}

\begin{table*}[ht]\small
  \begin{center}
    \caption{Computational complexity in contrast to best MPJPE on Human3.6M. FLOPs are reported for a single forward pass of the uplifting model. We also report the poses per second (\emph{PPS}) for a video frame rate of $50$~Hz on an NVIDIA 1080Ti.}
    \label{tab:complexity}
    \begin{tabular}{l|rrccc}
      \hline\noalign{\smallskip}
          Method&\# Params &FLOPs$\downarrow$&PPS (w/o CPN) $\uparrow$ &PPS (w/ CPN) $\uparrow$& MPJPE (mm)$\downarrow$ \\
          \noalign{\smallskip}
          \hline
          \noalign{\smallskip}
          Strided Transformer~\cite{li2022exploiting} ($N$ = 351) & 4.34 M & 2142 M & 208 & 32 & \textbf{43.7}  \\ 
      Pose Former~\cite{zheng20213d} ($N$ = 81) & 9.60 M & 1358 M & 248 & 33 & 44.3 \\
      \hl{Ours ($N$ = 81), $s_\inp$ = 10, $s_\outp$ = 2, + PT}  & 10.36 M  & \textbf{543 M} & 334 & 179 & 45.5  \\ 
          Ours ($N$ = 351), $s_\inp$ = 20, $s_\outp$ = 5, + PT  & 10.39 M  & 966 M & \textbf{827} & \textbf{399} & 45.0  \\ 
          \noalign{\smallskip}
          \hline
    \end{tabular}
  \end{center}
\end{table*}

\paragraph{Sparse Input Sequences}
Table~\ref{tab:sparse_results} shows results on Human3.6M with varying input strides $s_\inp$.
For Pose Former~\cite{zheng20213d} and Strided Transformer~\cite{li2022exploiting}, where $s_\outp = s_\inp$, we adopt bilinear interpolation between output poses at adjacent key-frames to obtain the full-rate 3D poses.
\hl{We evaluate our model with bilinear ($s_\outp = s_\inp$) and learned ($s_\outp = 2$) upsampling.}
At a moderate input stride of $s_\inp = 4$, we observe no difference in prediction quality between full frame rate poses and key-frame poses for all three architectures.
With increasing input stride, Strided Transformer results deteriorate notably in both key-frame and all-frame performance.
It shows that this architecture is only suitable for long and dense input sequences.
Pose Former shows more stable key-frame results, but the full frame rate predictions increasingly suffer from bilinear interpolation.
\hl{Our architecture, as a deeper combination of the former two, achieves lower spatial precision loss on key-frames with increasing input stride.
This advantage carries over to full frame rate results, but pure bilinear interpolation stays a limiting factor for high input strides ($s_\inp = 20$).
Finally, our explicit Transformer-based upsampling leads to a notably smaller gap between key-frame and all-frame performance on all metrics.}
It is better suited for temporally consistent, full frame rate 3D HPE on sparse input sequences.
At the same time, we have a single flexible model that supports different input rates of 2D poses.
Existing methods, including Pose Former and Strided Transformer, require a separate model for each input rate.

\paragraph{MoCap Pre-Training}

\begin{table}[t]\small
  \begin{center}
    \caption{Results on Human3.6M with $N=81$, with and without pre-training (\emph{PT}) on AMASS. 
    }
    \label{tab:amass_results}
    \resizebox{\columnwidth}{!}{
      \begin{tabular}{l|c|c|c}
        \hline\noalign{\smallskip}
        &&\multicolumn{2}{c}{MPJPE / N-MPJPE / P-MPJPE $\downarrow$}\\
        Method&\#Params&w/o PT&w/ PT \\
        \noalign{\smallskip}
        \hline
        \noalign{\smallskip}
        Strided Transformer~\cite{li2022exploiting}& $\,\,$4.06 M & 48.1 / 46.6 / 37.7 & 47.7 / 46.2 / 37.5 \\ 
        Pose Former~\cite{zheng20213d}&$\,\,$9.60 M &\textbf{47.4} / 46.0 / 37.4& 46.0 / 44.5 / 36.1 \\ 
        Ours, $s_\inp$ = $s_\outp$ = 2&10.36 M & 47.5 / \textbf{45.4} / \textbf{36.8} & \textbf{45.7} / \textbf{44.2} / \textbf{35.8}  \\ 
        \noalign{\smallskip}
        \hline
      \end{tabular}
    }
  \end{center}
\end{table}

Table~\ref{tab:amass_results} depicts results on Human3.6M, with and without MoCap pre-training on AMASS.
In this experiment we assume dense input sequences.
We compare the direct influence of pre-training on the different Transformer-based architectures.
We observe that all three uplifting Transformer architectures can improve on the target dataset with additional pre-training.
Strided Transformer, with its much lower network capacity, shows only marginal gains.
Our architecture achieves a similar MPJPE on pure Human3.6M training compared to Pose Former, but improves the most from pre-training. This shows the benefits of the deeper architecture at an otherwise comparable number of parameters.

\paragraph{Computational Complexity}

Finally, we compare the computational complexity of our method in Table~\ref{tab:complexity}.
When looking at the uplifting alone, a single forward pass \hl{with $N=351$} requires similar FLOPs compared to Pose Former and its much smaller sequence length.
Our uplifting model requires a forward pass for every $s_\outp$-th frame only, however.
Since the architecture is deeper and thus has more sequential computations, we achieve an uplifting speed-up of roughly $\times 4$ with $s_\outp=5$.
When we factor in the computational complexity of the a-priori 2D pose estimation, the true gain in efficiency becomes apparent (see also Figure~\ref{fig:flops}).
Using CPN to generate input 2D poses, Pose Former and Strided Transformer can no longer meet real-time throughput.
Since our model only requires input poses for every $s_\inp$-th frame, we observe a total speed-up of $\times 12$ for $s_\inp=20$, such that $50$~Hz videos can easily be processed with real-time throughput on a mid-range consumer GPU.
This estimate in speed-up is still rather pessimistic, since we measure CPN inference time for tight image crops.
The overhead for per-frame person detection is not included here.
The more 2D pose estimation dominates the total complexity, the more the speed-up factor will converge towards $s_\inp$.

\section{Conclusion}
\label{sec:conclusion}
We presented a Transformer-based 2D-to-3D pose uplifting model for efficient 3D human pose estimation in videos.
It uses building blocks from existing uplifting Transformers to form a deeper architecture that operates on sparse input sequences.
With a joint uplifting and upsampling Transformer module, the sparse 2D poses are
translated into dense 3D pose predictions.
This reduces the overall complexity but allows for temporally consistent 3D poses at the target frame rate of the video.
At the same time, we can train and deploy a single flexible model that can operate on different input frame rates.
The adverse effects of sparse input sequences can be greatly reduced with pre-training on large-scale motion capture archives. Our experiments reveal a speed-up by factor $12$ while still being competitive in spatial precision with current state-of-the art methods.

\newpage
\twocolumn[
    \centering
    \Large
    \bf
    Supplementary Material
    \vskip .5em
    \vspace*{12pt}
]

\setcounter{section}{0}
\renewcommand\thesection{\Alph{section}}

In the following, we provide additional experiments and qualitative results to further validate our design choices. This includes experiments on upsampling token attention, multi-stride training, within-batch augmentation and architecture components. We additionally discuss limitations of our method and areas of future work.

\section{Multi-Stride Training}
\label{sec:multi_stride_training}

By default, all our models are trained on multiple input strides $s_\inp$ simultaneously. We first take a closer look at the effects of multi-stride training.

\subsection{Single- vs. Multi-Stride Models}
\label{sec:single_vs_multi}

Multi-stride training has the benefit of a more flexible model that can operate on different input sample rates of 2D poses.
We evaluate, whether this flexibility comes at the cost of reduced spatial precision.
Table~\ref{tab:single_vs_multi_stride} compares a multi-stride model to separate single-stride models that are trained on one specific input stride $s_\inp = s_\outp$ each.
The results on Human3.6M~\cite{ionescu14} show that multi-stride training even improves 3D pose estimates.
For small input strides, single and multi-stride models are on par and show no real advantage of either training mode regarding spatial accuracy.
With increasing $s_\inp$, multi-stride training can consistently outperform the single-stride counterpart, on key-frame poses as well as all-frame results.
Thus, when aiming for very efficient operation with long input strides, multi-stride training leads to better uplifting and upsampling in 3D output space.

\begin{table}[t]\small
  \begin{center}
    \caption{Comparison of single- (\emph{SS}) and multi-stride (\emph{MS}) training on Human3.6M. All models are trained with $N=81$ and $s_\outp =2$. The MS models are trained on all input strides $s_\inp \in \lbrace 4,10,20 \rbrace$ simultaneously. By default, models use deferred upsampling token attention (\emph{DUTA}) within the temporal Transformer. Results are reported for poses on key-frames as well as all frames at $50$~Hz.}
    \label{tab:single_vs_multi_stride}
    \resizebox{\columnwidth}{!}{
      \begin{tabular}{l|c|c|c}
        \hline\noalign{\smallskip}
        &&\multicolumn{2}{c}{MPJPE / N-MPJPE / P-MPJPE $\downarrow$}\\
        &$s_{\inp}$&Key-frames&All frames \\
        \noalign{\smallskip}
        \hline
        \noalign{\smallskip}
        SS, w/o DUTA& 4  & 47.9 / \textbf{45.8} / \textbf{37.1}& 47.9 / \textbf{45.8} / \textbf{37.1} \\
        SS & 4 & 47.9 / \textbf{45.8} / \textbf{37.1}  & 47.9 / \textbf{45.8} / \textbf{37.1} \\
        MS, w/o DUTA& 4 & 59.7 / 54.0 / 43.8 & 52.7 / 49.0 / 39.6 \\
        MS & 4  & \textbf{47.6} / 46.0 / 37.3 & \textbf{47.4} / \textbf{45.8} / \textbf{37.1} \\
        \noalign{\smallskip}
        \hline
        \noalign{\smallskip}
        SS, w/o DUTA& 10  & 48.0 / 46.1 / 37.4 & 48.2 / 46.3 / 37.6 \\
        SS & 10  & 47.8 / 45.9 / \textbf{37.1} & 48.0 / \textbf{46.1} / \textbf{37.3} \\
        MS, w/o DUTA& 10 & 49.9 / 47.7 / 38.6 & 48.2 / 46.3 / 37.5 \\
        MS & 10  & \textbf{47.5} / \textbf{45.8} / \textbf{37.1} & \textbf{47.9} / \textbf{46.1} / 37.4 \\
        \noalign{\smallskip}
        \hline
        \noalign{\smallskip}
        SS, w/o DUTA & 20  & 49.3 / 47.1 / 38.1 & 50.6 / 48.4 / 39.3 \\
        SS & 20 & 50.1 / 47.4 / 38.4 & 51.5 / 48.8 / 39.6  \\
        MS, w/o DUTA& 20  & 49.4 / 47.4 / 38.5 & 52.1 / 49.8 / 40.5 \\
        MS & 20  & \textbf{48.2} / \textbf{46.4} / \textbf{37.6} & \textbf{49.9} / \textbf{48.1} / \textbf{39.2} \\
        \noalign{\smallskip}
        \hline
      \end{tabular}
    }
  \end{center}
  \vspace{-0.4cm}
\end{table}

\begin{figure*}[t]
  \begin{centering}
    \includegraphics[width=0.93\linewidth]{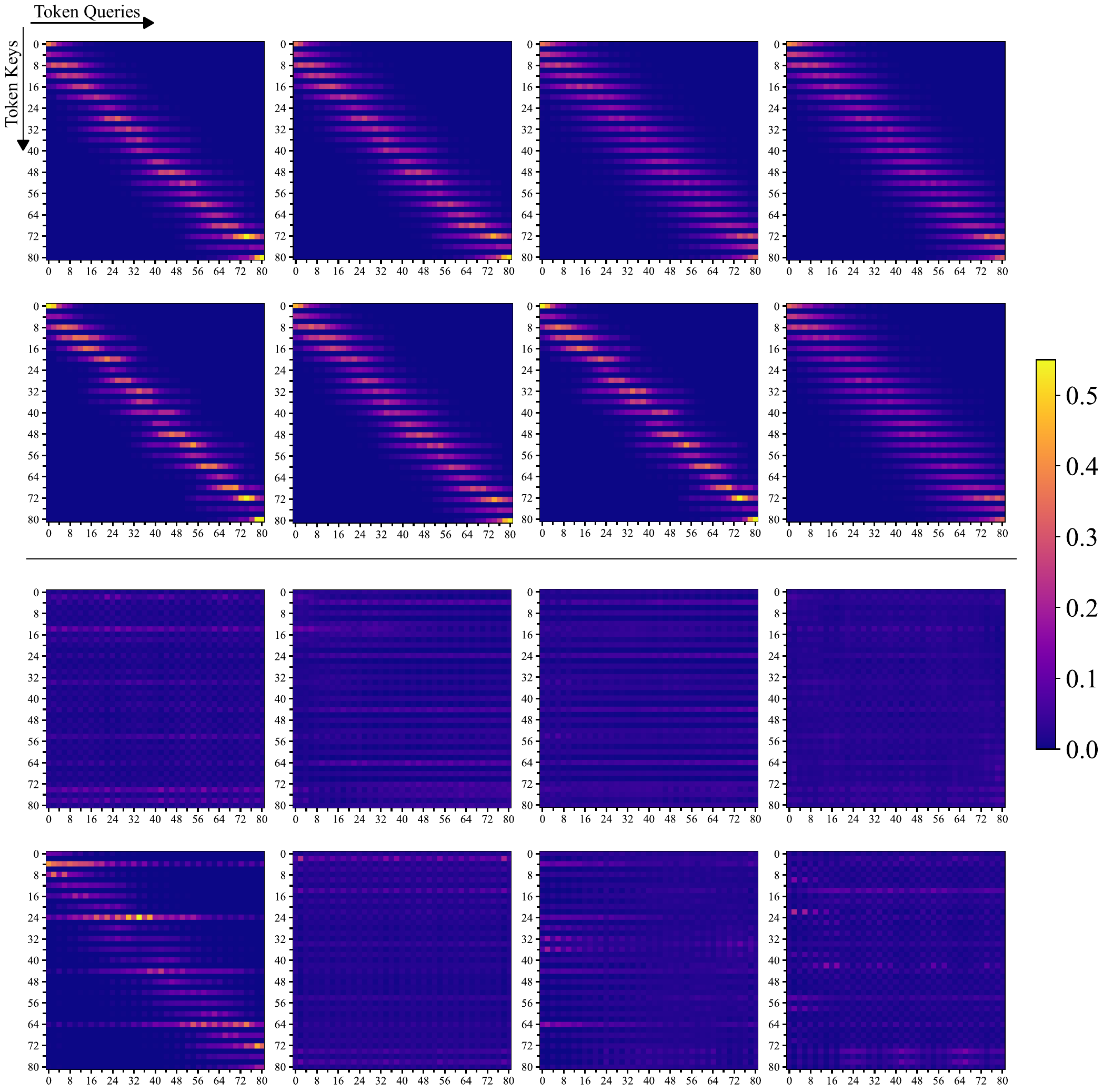}
    \caption{Token attention of the eight MHA heads in the first temporal Transformer block. Top: With DUTA. Bottom: Without DUTA. The corresponding models are trained with $N=81$, $s_\outp = 2$ and $s_\inp \in \lbrace 4, 10, 20 \rbrace$. The examples are generated with $s_\inp = 4$, \ie with a ratio of pose and upsampling tokens of one.}
    \label{fig:attention}
  \end{centering}
\end{figure*}

\subsection{Upsampling Token Attention}
\label{sec:duta}
At the beginning of the very first temporal Transformer block, none of the upsampling tokens carry any input-related information.
They are only conditioned on their relative frame index.
Therefore, any attention to the upsampling tokens will not lead to meaningful information exchange or gain.
In contrast, there is even the risk of deteriorating the information carried by the actual pose tokens.
To counter this effect, we only allow attention to upsampling tokens from the second temporal Transformer block onward.
At this stage, all tokens carry input-related information to some degree.
We refer to this as deferred upsampling token attention (DUTA).
Table~\ref{tab:single_vs_multi_stride} shows results for our multi-stride model on Human3.6M, with and without DUTA.
The results clearly show the necessity for DUTA, as it outperforms the vanilla variant with unconstrained cross-token attention on all inputs strides and metrics.
The most notable difference occurs when evaluating with $s_\inp = 4$.
In this setting, training without DUTA leads to worse key-frame performance compared to all-frame results.
This shows that the pose token representation heavily suffers from unconstrained attention within the first temporal Transformer block.
The negative effects are less severe for larger $s_\inp$, but there is still a clear performance gap to the DUTA variant.
Thus, delaying the full cross-token attention is a crucial design choice for stable results over different input strides.
Table~\ref{tab:single_vs_multi_stride} also shows the influence of DUTA on single-stride models. For single-stride training, the ratio between pose and upsampling tokens stays the same for all training examples. The results reveal no clear advantage or disadvantage when training with DUTA in this setting.
This shows that DUTA is only required for a variable ratio of pose and upsampling tokens with multi-stride training.
Figure~\ref{fig:attention} depicts exemplary token attention within the first temporal Transformer block of a multi-stride model. With DUTA, temporal attention shows reasonable token-local information aggregation as often seen in the early stages of a temporal Transformer.
Without DUTA, temporal attention is uniformly spread over a distant subset of pose tokens as well as intermediate upsampling tokens.
This seems to greatly hinder proper information exchange over the temporal sequence.


\section{Within-Batch Augmentation}
\label{sec:wba}

\begin{table}[t]\small
  \begin{center}
    \caption{Effects of within-batch augmentation (\emph{WBA}) on Human3.6M with $N=351$, $s_\outp = 5$, $s_\inp \in \lbrace 5, 10, 20 \rbrace$ and batch size $512$. Results are shown with and without pre-training on motion capture sequences from AMASS.}
    \label{tab:wba}
    \resizebox{\columnwidth}{!}{
      \begin{tabular}{c|c|c|c}
        \hline\noalign{\smallskip}
        &&\multicolumn{2}{c}{MPJPE / N-MPJPE / P-MPJPE $\downarrow$}\\
        WBA&$s_\inp$&w/o PT&w/ PT \\
        \noalign{\smallskip}
        \hline
        \noalign{\smallskip}
        \xmark&5 & 46.0 / 44.4 / 36.5  & 43.5 / 42.1 / 34.7   \\ 
        \cmark&5 & \textbf{45.7} / \textbf{44.3} / \textbf{36.4} & \textbf{42.6} / \textbf{41.5} / \textbf{34.2}  \\ 
        \noalign{\smallskip}
        \hline
        \noalign{\smallskip}
        \xmark&20 & 48.2 / 46.7 / 38.6  & 45.7 / 44.4 / 36.8   \\ 
        \cmark&20 & \textbf{47.8} / \textbf{46.4} / \textbf{38.4} & \textbf{45.0} / \textbf{44.0} / \textbf{36.3}  \\ 
        \noalign{\smallskip}
        \hline
      \end{tabular}
    }
  \end{center}
    \vspace{-0.4cm}
\end{table}

Next, we evaluate the benefits of within-batch augmentation (WBA).
With WBA, each mini-batch contains the flip-augmented and non-augmented version of a training example.
This promotes invariance to horizontal flipping within each weight update.
Table~\ref{tab:wba} shows the results on Human3.6M. WBA leads to a slight performance gain in all metrics, independent of the input stride.
Note that this benefit comes at no additional cost during training, since all models in this comparison use a fixed batch size of 512.
The result shows that the benefits of WBA outweigh the effectively reduced variability within each mini-batch. We also evaluate with additional pre-training on AMASS.
This setting reveals a significant boost through WBA, with a reduction of up to $0.9$~mm in MPJPE.
We observe that WBA leads to slightly slower convergence during pre-training, but far better validation results.
This advantage is then translated over to the fine-tuning on Human3.6M.
We also observe similar benefits when experimenting with Pose Former and, to a lesser extent, Strided Transformer.

\begin{figure*}[t]
  \begin{centering}
    \includegraphics[width=0.85\linewidth]{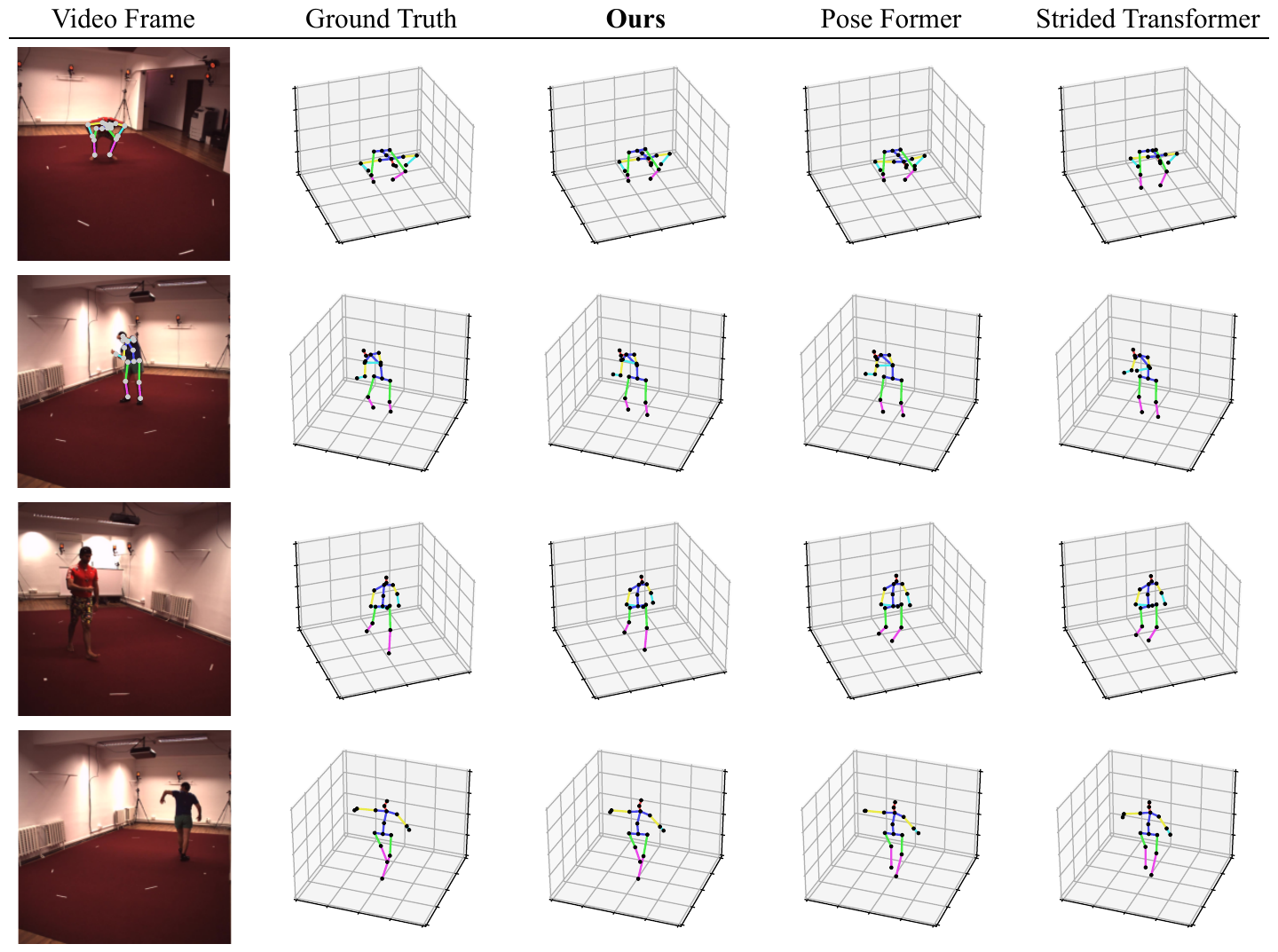}
    \caption{Qualitative examples on Human3.6M, with $N=81$ and $s_\inp = 20$. We compare our method, Pose Former~\cite{zheng20213d} and Strided Transformer~\cite{li2022exploiting}. Top: Examples on key-frames. Bottom: Examples on non-key-frames.}
    \label{fig:qualitative_ours_pf_st}
  \end{centering}
\end{figure*}

\section{\hl{Architecture Components}}
\label{sec:architecture}

We also evaluate the individual influence of the three main components of our Transformer architecture: 
The joint-wise spatial Transformer, the pose-wise temporal Transformer  and the strided Transformer. 
Table~\ref{tab:architecture} compares our full architecture to variants where one component is removed at a time. 
Starting with the temporal Transformer, this component is the most crucial part of our and related architectures~\cite{li2022exploiting, shan22}. 
Removing this block disables repeated self-attention across the entire sequence of pose and upsampling tokens. Additionally, it impedes the full sequence loss $\mathcal{L}_{\text{seq}}$. In combination, the results show that the temporal Transformer is a strict requirement for our architecture to operate properly.
Removing the spatial Transformer is less impactful, but we observe a clear drop in precision across all input strides. 
Thus, dedicating a separate Transformer to generate an initial pose representation is beneficial, especially when input 2D poses are temporally sparse. 
Finally, the strided Transformer has the lowest impact compared to the other two components, but its removal still leads to an increase in MPJPE by $0.6$ - $0.9$~mm. It acts as a refinement component via the center frame loss $\mathcal{L}_{\text{center}}$ and is again most helpful for large input strides. 
Due to the internal temporal striding, it is computationally less expensive compared to the full temporal Transformer blocks and therefore a valid addition to our architecture.

\begin{table}[t]\small
  \begin{center}
    \caption{MPJPE (mm) on Human3.6M with $N=81$, $s_\outp =2$ and $s_\inp \in \lbrace 4, 10, 20 \rbrace$. We compare variants of our architecture with either the spatial Transformer (SPT), the temporal Transformer (TT) or the strided Transformer (ST) removed. The MPJPE is reported relative to the full architecture results.}
    \label{tab:architecture}
    \resizebox{\columnwidth}{!}{
      \begin{tabular}{ccc|ccc}
        \hline\noalign{\smallskip}
        SPT&TT&ST&$s_\inp=4$&$s_\inp=10$&$s_\inp=20$ \\
        \noalign{\smallskip}
        \hline
        \noalign{\smallskip}
        \cmark & \cmark & \cmark & \textbf{47.4} & \textbf{47.9} & \textbf{49.9} \\ 
        \noalign{\smallskip}
        \hline
        \noalign{\smallskip}
         & \cmark & \cmark & $+1.2$ & $+1.2$ & $+1.5$ \\ 
        \cmark &  & \cmark & $+4.2$ & $+4.7$ & $+5.9$ \\ 
        \cmark & \cmark &  & $+0.6$ & $+0.7$ & $+0.9$ \\ 
        
        \noalign{\smallskip}
        \hline
      \end{tabular}
    }
  \end{center}
    \vspace{-0.4cm}
\end{table}

\section{Qualitative Examples}
\label{sec:qualitative}

Figure~\ref{fig:qualitative_ours_pf_st} depicts qualitative examples on Human3.6M with our method, Pose Former and Strided Transformer at an input stride of $s_\inp=20$.
In comparison to Strided Transformer, our method typically leads to more precise 3D estimates on key-frames an non-key-frames.
Pose Former is more robust to sparse input sequences, but our method still leads to better results on human motion at non-key-frames, \eg during walking motion.
Figure~\ref{fig:qualitative_ours} depicts additional examples from our best models and 2D input poses at $2.5$~Hz.
We observe plausible human motion even on difficult examples within Human3.6M and MPI-INF-3DHP.
\hl{The examples in rows three and six depict failure cases, which we discuss in detail next.}

\section{\hl{Error Modes and Limitations}}
\label{sec:limitations}

\begin{figure}[t]
  \begin{centering}
    \includegraphics[width=0.99\linewidth]{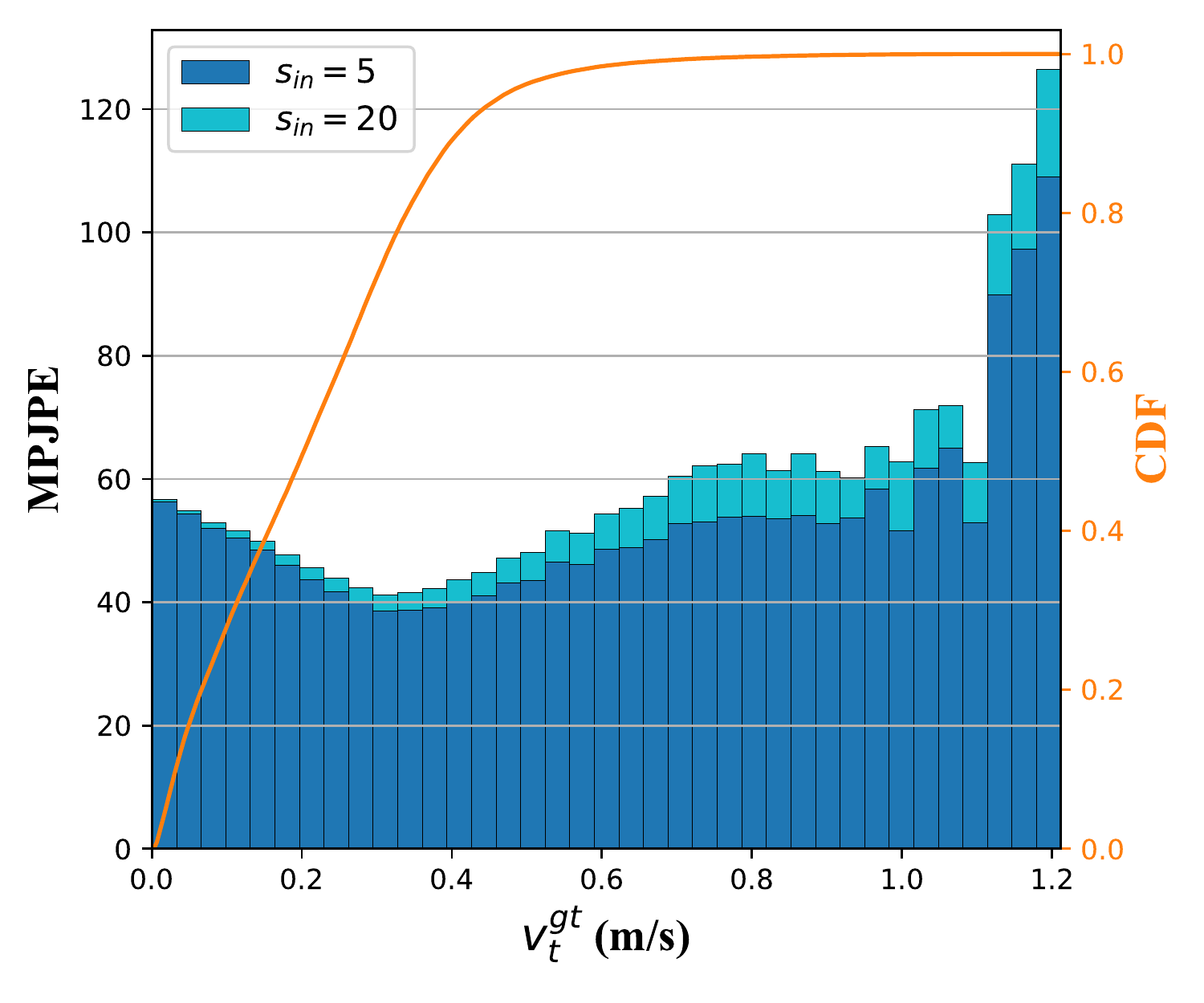}
    \caption{The MPJPE on Human3.6M in contrast to the ground truth pose velocity $v_t^{\text{gt}}$. The velocities are discretized into equally sized intervals. We additionally show the cumulative distribution (CDF) over the velocities in the dataset.}
    \label{fig:mpjpe_velocity}
  \end{centering}
\end{figure}

\begin{figure*}[t]
  \begin{centering}
    \includegraphics[width=0.99\linewidth]{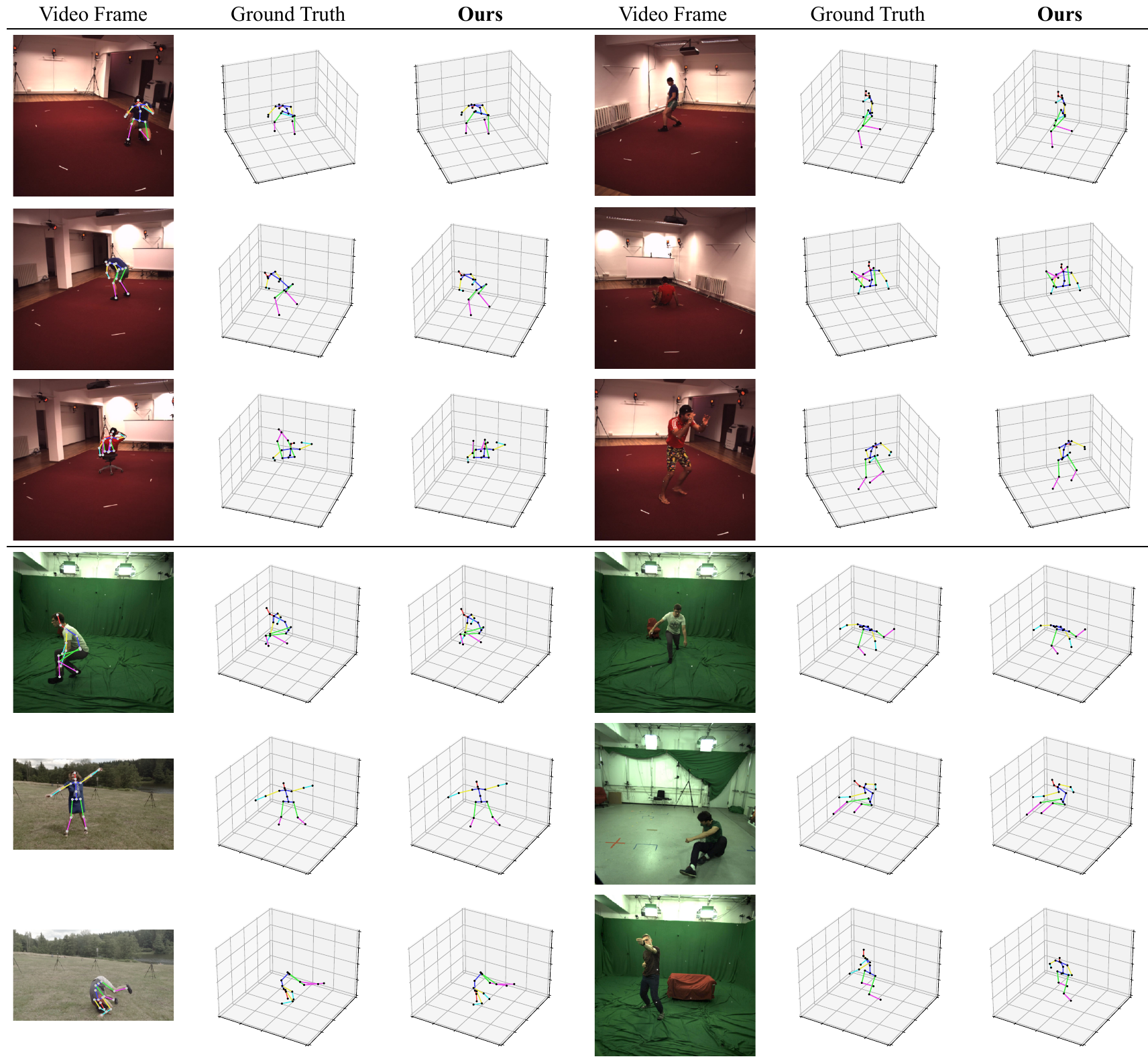}
    \caption{Qualitative examples on Human3.6M (top) and MPI-INF-3DHP~\cite{mehta17} (bottom). The results are generated with our best models and 2D poses at $2.5$~Hz. The left column shows results on key-frames, the right column on non-key-frames. Failure cases are depicted in rows three and six.}
    \label{fig:qualitative_ours}
  \end{centering}
\end{figure*}

We observe two main error modes for our proposed method.
The first cause of erroneous 3D pose estimates are missdetections within the 2D pose estimates.
Figure~\ref{fig:qualitative_ours} (rows three, left) depicts such an example, where the estimated 2D locations of leg joints suffer from self-occlusion.
Note that the dependence on high quality 2D poses is common to all uplifting methods~\cite{pavllo19, wang20, li2022exploiting}.
Therefore, we see this error mode as a limitation of single-frame 2D HPE and the 2D-to-3D pose uplifting approach in general.

The second error mode is more unique to our method.
Figure~\ref{fig:qualitative_ours} (row six, right) depicts an example of a person performing boxing punches in quick succession.
We observe that some of the punches are not reconstructed within the estimated 3D pose sequence.
Since we utilize input 2D poses at only 2.5~Hz for this example, some of the punches occur so fast that they are completely missing from the subsampled input sequence as well.
Consequently, our model is not able to reconstruct the full motion.

In order to analyze the dependency between 2D pose subsampling and fast body motion, we define the average root-relative velocity $v_t$ of a pose $P_t$ as 
\begin{equation}
  \label{eq:velocity}
  v_t = \frac{1}{J} \sum_{j=1}^{J} \lVert \left( P_{t,j} - P_{t,r} \right) - \left( P_{t-1,j} - P_{t-1,r} \right) \rVert_2,
\end{equation}
where again the pelvis is used as the root joint $r$.
We use the relative velocity, since we focus on the speed of within-body movement.
We want to measure fast movement of \eg the arms or legs independent of the person standing in place or walking.
Figure~\ref{fig:mpjpe_velocity} shows the MPJPE on Human3.6M in contrast to the ground truth velocity $v_t^{\text{gt}}$ (reported in m/s).
We divide the range of observable velocities into equidistant intervals and report the MPJPE for all estimated 3D poses within an interval.
The results are depicted for the same model under two different settings:
a moderate input stride of $s_\inp = 5$ for high spatial precision and a long input stride of $s_\inp = 20$ for best efficiency.
Under no or very small movement ($<0.2$~m/s), both settings perform equally.
Due to the limited motion, the temporal component of the input sequence does not offer additional information, no matter the input stride.
For movement in the range of $0.2$ - $0.4$~m/s (\eg walking), both settings show rather stable results, with only minor losses in precision for $s_\inp = 20$.
Most actions within Human3.6M fall into this range of relative pose velocity.
Only for considerably faster movement, the results of both settings diverge.
While $s_\inp = 5$ (\ie 2D poses at 10~Hz) stays relatively stable around $50$~mm MPJPE, our fastest setting with $s_\inp = 20$ shows increasing difficulties in reconstructing the true pose sequence in 3D space.
This reveals the main limitation of our method:
The choice of efficiency, which is mainly governed by $s_\inp$, must not only fit potential hardware and runtime requirements, but also the range of expected movement speed.
While our most efficient setting with $s_\inp = 20$ is suitable for regular movement in daily life, it will not fit applications in \eg sporting activities.

\begin{table}[t]\small
  \begin{center}
    \caption{Computational complexity and best MPJPE (mm) on Human3.6M with MoCap pre-training. FLOPs are reported for a single forward pass of the uplifting model. We also report the poses per second (\emph{PPS}) for a video frame rate of $50$~Hz on an NVIDIA 1080Ti.}
    \label{tab:complexity_extended}
    \resizebox{0.99\linewidth}{!}{
      \begin{tabular}{c|c|c|rccc}
        \hline\noalign{\smallskip}
        \thead{$N$}& \thead{$s_\outp$} & \thead{$s_\inp$} & \thead{FLOPs$\downarrow$}&\thead{PPS$\uparrow$\\(w/o CPN)}&\thead{PPS$\uparrow$\\(w/ CPN)}& \thead{MPJPE$\downarrow$} \\
        \noalign{\smallskip}
        \hline
        \noalign{\smallskip}
        \multirow{3}{*}{81} & \multirow{3}{*}{2} & 4    & 564 M & 326 & 105 & 44.8  \\ 
                   &  & 10  & 543 M & 334 & 179 & 45.5  \\ 
                   &  & 20   & \textbf{535 M} & 337 & 234 & 47.9  \\ 
        \noalign{\smallskip}
        \hline
        \noalign{\smallskip}
        \multirow{3}{*}{351} & \multirow{3}{*}{5} & 5    & 1062 M & 704 & 151 & \textbf{42.6}  \\ 
                   &  & 10  & 999 M & 759 & 255 & 43.1  \\ 
                   &  & 20 & 966 M & \textbf{827} & \textbf{399} & 45.0  \\ 
        \noalign{\smallskip}
        \hline
      \end{tabular}
    }
  \end{center}
\end{table}

\section{\hl{Adaptive Input Stride}}
\label{sec:adaptiuve input stride}

Finally, we discuss further potential of our method that is yet to be exploited. 
One of the main advantages of our method is that a single instance of our uplifting model (\ie a single set of model parameters) can support different input strides. 
Thus, a single model can be operated with different computational complexity and processing rates (see Table~\ref{tab:complexity_extended} for extended runtime and complexity results). 
For all experiments in this paper, the input stride is kept constant throughout the processing of an entire video. 
This is no strict requirement though. 
A change in input stride only affects how many video frames are used for 2D pose estimation and subsequent pose token generation within the spatial Transformer.
No other reconfiguration of our model is required. 
Thus, the input stride can be changed online while processing a video stream. 
This enables hardware-limited devices to dynamically adapt to currently available shared resources like memory or computational units (CPUs, GPUs, TPUs).

\begin{figure}[t]
  \begin{centering}
    \includegraphics[width=0.99\linewidth]{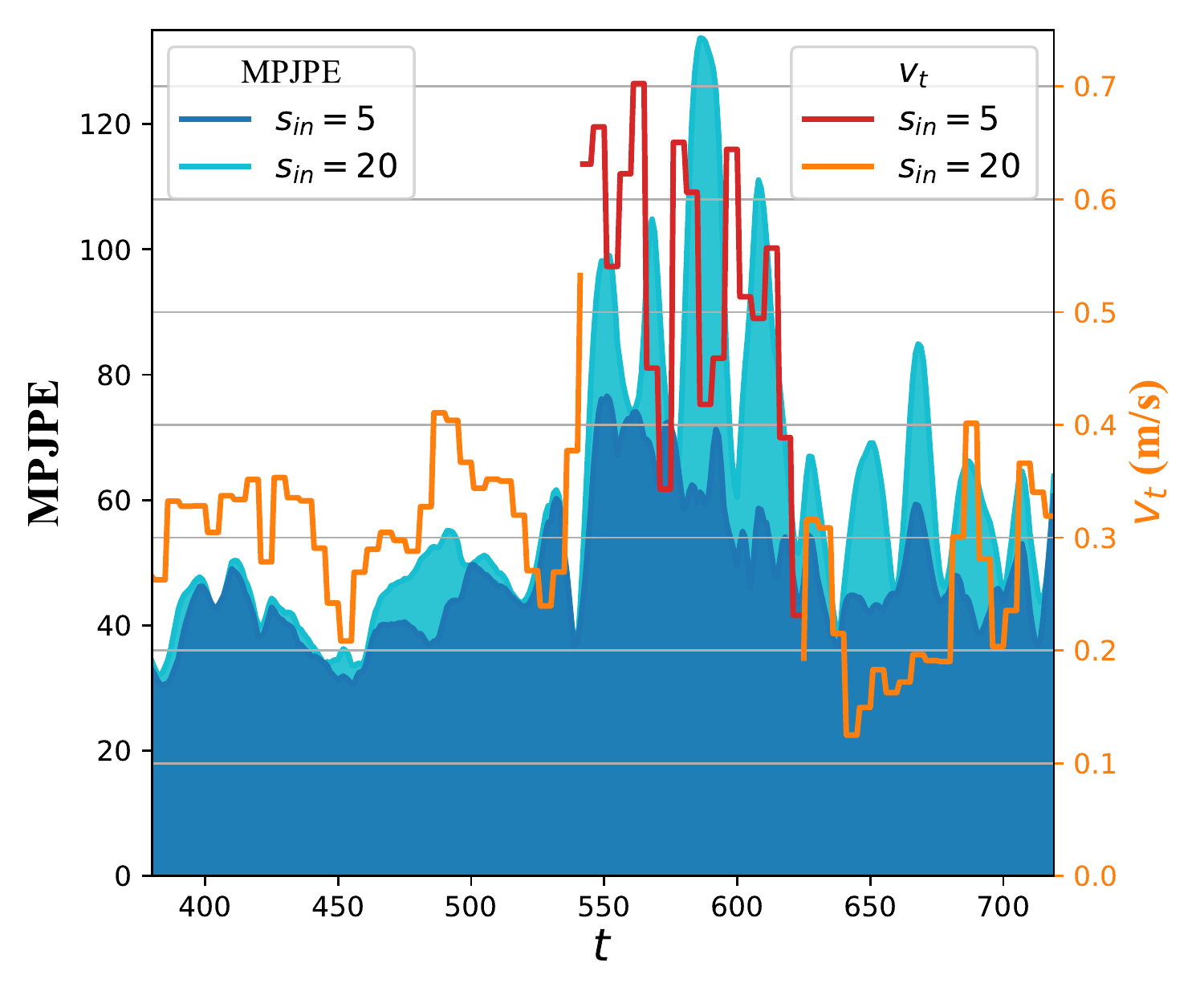}
    \caption{MPJPE on an exemplary "WalkDog" video from Human3.6M, with $N=351$ and $s_\outp=5$. We switch our model from $s_\inp = 20$ to $s_\inp = 5$ for video sections where the observed relative pose velocity $v_t$ surpasses $0.5$~m/s (red).}
    \label{fig:adaptive_mode}
  \end{centering}
\end{figure}

A second use case of variable input strides is the adaption to the occurring human motion. 
Based on the discussion about movement speed in Section~\ref{sec:limitations}, we can process a video with long input stride by default, and only switch to a shorter input stride for increased precision when observing fast body movement.
Figure~\ref{fig:adaptive_mode} represents an exemplary Human3.6M video where a short sequence of running occurs. 
By thresholding the pose velocity $v_t$ from the 3D pose estimates (orange), \eg with $0.5$~m/s, we can identify this video section and switch from $s_\inp =20$ to $s_\inp =5$. 
Only when the velocity (red) drops below the threshold for a fixed number of frames, we switch back to the more efficient input stride.
This way we avoid the otherwise failed 3D pose estimation with an MPJPE $>80$~mm.
Note that the relative velocity is only one of many possible statistics for identifying difficult video sections. 
We leave the development and evaluation of such statistics as a research direction for future work.

{\small
\bibliographystyle{ieee_fullname}
\bibliography{bibliography}
}

\end{document}